\documentclass{article}
\usepackage{graphicx} 

\usepackage{amssymb}
\usepackage{amsmath}
\usepackage{mathtools}
\usepackage[ruled,linesnumbered]{algorithm2e}
\usepackage{multirow}
\usepackage{xcolor}
\usepackage{lscape}
\usepackage{enumitem}
\usepackage{subcaption}
\usepackage{hyperref}

\def\R{\mathbb{R}}

\title{In-Context Operator Learning with Data Prompts for Differential Equation Problems}
\author{Liu Yang, Siting Liu, Tingwei Meng, Stanley J. Osher\footnote{Corresponding Author: sjo@math.ucla.edu, the second and third authors contributed equally. \\ 
\textcolor{red}{This is an outdated preprint. Please refer to the updated version published in PNAS: \url{www.pnas.org/doi/10.1073/pnas.2310142120}} \\
See code in \url{https://github.com/LiuYangMage/in-context-operator-networks}
}}
\date{Department of Mathematics, UCLA, Los Angeles, CA 90095, USA}

\begin{document}

\maketitle

\begin{abstract}
This paper introduces a new neural-network-based approach, namely In-Context Operator Networks (ICON), to simultaneously learn operators from the prompted data and apply it to new questions during the inference stage, without any weight update. Existing methods are limited to using a neural network to approximate a specific equation solution or a specific operator, requiring retraining when switching to a new problem with different equations. By training a single neural network as an operator learner, we can not only get rid of retraining (even fine-tuning) the neural network for new problems, but also leverage the commonalities shared across operators so that only a few demos in the prompt are needed when learning a new operator. Our numerical results show the neural network's capability as a few-shot operator learner for a diversified type of differential equation problems, including forward and inverse problems of ordinary differential equations (ODEs), partial differential equations (PDEs), and mean-field control (MFC) problems, and also show that it can generalize its learning capability to operators beyond the training distribution.

\end{abstract}

\section{Introduction}

The development of neural networks has brought a significant impact on solving differential equation problems. We refer the readers to~\cite{karniadakis2021physics} for the recent advancement in this topic.

One typical approach aims to directly approximate the solution given a specific problem. The Physics-Informed Neural Networks (PINNs)~\cite{Raissi2019PINN} propose a neural network method for solving both forward and inverse problems by integrating both data and differential equations in the loss function. Deep Galerkin Method (DGM)~\cite{Sirignano2018DGM} imposes constraints on the neural networks to satisfy the prescribed differential equations and boundary conditions. Deep Ritz Method (DRM)~\cite{E2018deepRitz} utilizes the variational form of PDEs and can be used for solving PDEs that can be transformed into equivalent energy minimization problems. Weak Adversarial Network (WAN)~\cite{Zang2020Weakadversarial} leverages the weak form of PDEs by parameterizing the weak solution and the test function as the primal and adversarial networks, respectively. \cite{Han2018Solving} proposed a deep learning method based on the stochastic representation of high-dimensional parabolic PDEs. \cite{Ruthotto2020machine} solves high-dimensional mean-field game problems by encoding both Lagrangian and Eulerian viewpoints in neural network parameterization. APAC-net~\cite{Lin2021Alternating} proposes a generative adversarial network style method that utilizes the primal-dual formulation for solving mean-field game problems.

Despite their success, the above methods are designed to solve problems with a specific differential equation. The neural network needs to be trained again when the terms in the equation or the initial/boundary conditions change.

Later, efforts have been made to approximate the solution operator for a differential equation with different parameters or initial/boundary conditions. 
PDE-Net~\cite{long2018pde} utilizes convolution kernels to learn differential operators, allowing it to unveil the evolution PDE model from data, and make forward predictions with the learned solution map. 
Deep Operator Network (DeepONet)~\cite{lu2021learning} designed a neural network architecture to approximate the solution operator which maps the parameters or the initial/boundary conditions to the solutions.
Fourier Neural Operator (FNO)~\cite{li2021fourier} utilizes the Fourier transform to learn the solution operator.

The above methods have successfully demonstrated the capability of neural networks in approximating solution operators. However, in these methods, ``one" neural network is limited to approximating ``one" operator. Even a minor change in the differential equation can cause a shift in the solution operator. For example, in the case of learning a solution operator mapping from the diffusion coefficient to the solution of a Poisson equation, the solution operator changes if the source term (which is not designed as a part of the operator input) changes, or a new term is introduced to the equation. Consequently, the neural network must be retrained.

We argue that there are commonalities shared across various solution operators. By using a single neural network with a single set of weights to learn various solution operators, we can not only get rid of retraining (even fine-tuning) the neural network, but also leverage such commonalities so that fewer data are needed when learning a new operator.

If we view learning one solution operator as one task, then we are now targeting solving multiple differential-equation-related tasks with a single neural network. Our expectation for this neural network goes beyond simply learning a specific operator. Rather, we expect it to acquire the ability to ``learn an operator from data'' and apply the newly learned operator to new problems.

Such ability to learn and apply new operators might be a very important part of artificial general intelligence (AGI). By observing the inputs and outputs of a physical system, a human could learn the underlying operator mapping inputs to outputs, and control the system according to their goals. For example, a motorcyclist can quickly adapt to a new motorcycle; a kayaker can quickly adapt to a new kayak or varying water conditions. If a human has expertise in both sports, they may be able to master jet skiing at their first few attempts. We expect a robot with AGI able to adapt to new environments and tasks, just as a human would.

The paradigm of ``learning to learn'', or meta-learning, has achieved great success in the recent development of artificial intelligence. In natural language processing (NLP), in-context learning introduced in GPT-2~\cite{radford2019language} and further scaled up in GPT-3~\cite{brown2020language} has demonstrated the capability of large language models (LLM) as few-shot learners. In-context learning gets rid of the limitations of the previous paradigm, i.e. pre-training plus fine-tuning, including (1) the need to fine-tune the neural network with a relatively large dataset for every new task, (2) the potential to overfit during fine-tuning which leads to poor out-of-distribution generalization, and (3) the lack of ability to seamlessly switch between or mix together multiple skills.

In this paper, we adapt the idea of in-context learning to learn operators for differential equations problems. 

\begin{figure}[ht]
\centering
\includegraphics[width=0.9\textwidth]{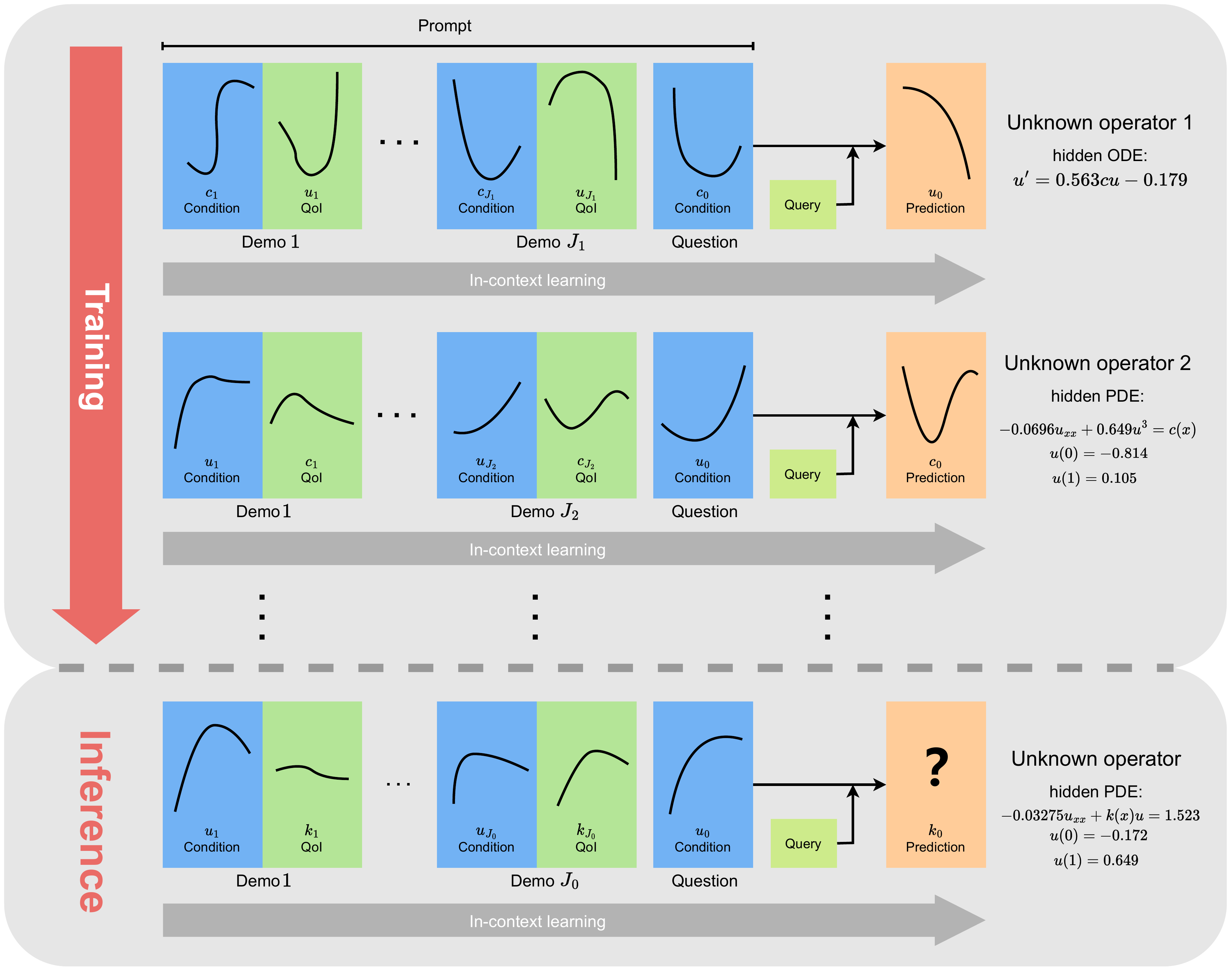}
\caption{Training and inference of In-Context Operator Networks (ICON). 
}
\label{fig:illustration_intro}
\end{figure}

We refer to the inputs of the operators as ``conditions", and the operator outputs as ``quantities of interest (QoIs)". A ``demo'' consists of one pair of condition and QoI. In the previous paradigm of operator learning~\cite{lu2021learning}~\cite{li2021fourier}, the neural network is trained on demos that share the same operator. During the inference stage, it takes a new condition as input and predicts the QoI corresponding to the learned operator. 
There are other non-neural-network methods in the similar paradigm, such as SINDy~\cite{brunton2016discovering} for solving inverse ODE problems via sparse regression over ODE parameters.
In this paper, during the inference stage, we instead have the trained neural network taking the demos and a new condition (namely ``question condition'') as input, and simultaneously completing the following two jobs: (1) learn the operator from demos, (2) apply the learned operator to the question condition and predict the corresponding QoI. We emphasize that there are no weight updates during the inference stage. We name the proposed method as In-Context Operator Networks, or ``ICON'' in short. The illustration of the training and inference of ICON is shown in Figure~\ref{fig:illustration_intro}.

The rest parts of the paper are organized as follows. In Section~\ref{sec:problem}, we introduce the problem setup of in-context operator learning. In Section~\ref{sec:method}, we present the detailed methodology of ICON. In Section~\ref{sec:experiments}, we present the experimental results, where we show the capability of ICON in learning operators from demos and applying to question conditions during the inference stage. 
In Section~\ref{sec:discussion}, we try to answer the question why a few demos are sufficient in learning the operator in the proposed method. In Section~\ref{sec:summary}, we conclude the paper and discuss the limitations and future work.
\section{Problem Setup}~\label{sec:problem}

In this section, we introduce the problem setup of in-context operator learning.

An operator is defined as a mapping that takes either a single input function or a tuple of input functions and produces an output function. In this paper, we refer to the inputs of the operators as the ``condition'', and the operator outputs as the ``quantities of interest (QoI)''.

Take a one-dimensional ODE problem $u'(t) = \alpha u(t) + \beta c(t) + \gamma$ as an example. Given the parameters $\alpha, \beta, \gamma\in\R$, there exists a solution operator that maps from the function $c\colon [0,T]\to \R$ and the initial condition $u(0)$, to $u \colon [0,T]\to \R$. In this case, $c\colon [0,T]\to \R$ and the initial condition $u(0)$ form the condition, and $u \colon [0,T]\to \R$ is the QoI. Note that while $u(0)$ is a number, we can still view it as a function on the domain $\{0\}$ to fit into the framework.  

In practical scenarios, it is often challenging to obtain an analytical representation of conditions and QoIs. Instead, we typically rely on observations or data collected from the system. To address this, we utilize a flexible and generalizable approach by representing these entities using key-value pairs, where keys are discrete function inputs, and values are the corresponding outputs of the function. Continuing with the example of the one-dimensional ODE problem introduced above, to represent the function $c\colon [0,T]\to\R$, we consider the discrete time instances as the keys, and the corresponding function values of $c$ as the associated values. we use the key $0$ and value $u(0)$ to represent the initial condition of $u$. It is important to note that the number of key-value pairs is arbitrary, the arrangement of keys is flexible, and they can vary across different functions.

The training data can be represented as $\{\{(\text{cond}_{i}^j, \text{QoI}_{i}^j)\}_{j=1}^{N_i}\}_{i=1}^M$, where each $i$ corresponds to a different operator. For a given $i$, $\{(\text{cond}_{i}^j, \text{QoI}_{i}^j)\}_{j=1}^{N_i}$ represents a set of $N_i$ condition-QoI pairs that share the same operator. In our setup, it is important to emphasize that the operators here are completely unknown, even in terms of the corresponding differential equation types. This aspect is aligned with many real-world scenarios where either the parameters of the governing equations are missing or the equation themselves need to be constructed from scratch.

During the inference stage, we are presented with pairs of conditions and QoIs, referred to as ``demos'', that also share an unknown operator. Additionally, we are given a condition called the ``question condition''. The objective is to predict the QoI corresponding to the question condition and the unknown operator. Note that the unknown operator in the inference stage may differ from the operators present in the training dataset, potentially even being out of distribution.

\section{Methodology}\label{sec:method}

In this section, we will introduce the method in detail. In Section~\ref{sec:prompt-query}, we show how to build the neural network inputs, including prompts and queries. In Section~\ref{sec:nn-architecture}, we illustrate the neural network architecture. We introduce the data preparation and training process in Section~\ref{sec:training}, and the inference process in Section~\ref{sec:inference}.

\subsection{Prompt and Query}\label{sec:prompt-query}
The model is expected to learn the operator from multiple demos, each consisting of a pair of condition and QoI, and apply it to the question condition, making predictions on the question QoI. As the question QoI is a function, it's also necessary to specify where the model should make evaluations, i.e., the keys for the question QoI, which is referred to as the ``queries'' (each query is a vector). We group the demos and question condition as ``prompts'', which together with the queries are the neural network inputs. The output of the neural network represents the prediction for the values of the question QoI, corresponding to the input queries.

Although alternative approaches exist, in this paper, we choose a simple method for constructing the prompts, wherein we concatenate the demos and the question condition to create a matrix representation. Each column of the matrix represents a key-value pair. Given that we will be using transformers, the arrangement of columns in the prompt will not affect the outcome. Therefore, in order to distinguish the key-value pairs from different conditions and QoIs, we concatenate the key and value in each column with an index column vector. Suppose the maximum capacity of demos is $J_{m}$, for simplicity, we use index vector $\textbf{e}_j$ for the condition in $j$-th demo, and $-\textbf{e}_j$ for the QoI in $j$-th demo, where $\textbf{e}_j$ is the one-hot column vector of size $J_{m}+1$ with the $j$-th component to be $1$. The index vector for the question condition is $\textbf{e}_{J_{m}+1}$. We remark that for a large $J_{m}$, a more compact representation, such as the trigonometric position embedding used in NLP tasks, can be applied. 

In order to cater to operators with varying numbers of input condition functions, and functions from different spaces, we restructure the keys in prompts and queries. Specifically, we assign the first row of the prompts/queries to indicate different function terms, the second row to denote temporal coordinates, the third row for the first spatial coordinate, and so forth. If certain entries are not required, we will populate them with zeros.

In table \ref{tab:sec2_demok_linearode}, we show the matrix representation of the $j$-th demo for the one-dimensional forward ODE problem aforementioned in Section~\ref{sec:problem}. The prompt is simply the concatenation of demos and the question condition along the row.

\begin{table}[!htp]\centering
\footnotesize
\begin{tabular}{ll|l|l}\hline
& & \multicolumn{1}{c|}{condition} & \multicolumn{1}{c}{QoI} \\\hline
\multicolumn{1}{c|}{}  &term &\multirow{5}{*}{$\begin{pmatrix} 0 & 0 & \dots & 0 & 1\\ t_1 & t_2 & \dots & t_{n_j-1} & 0\\ 0 & 0 & \dots & 0 & 0\\ c(t_1) & c(t_2) & \dots & c(t_{n_j-1}) & u(0)\\ \textbf{e}_j & \textbf{e}_j & \dots & \textbf{e}_j& \textbf{e}_j \end{pmatrix}$ } &\multirow{5}{*}{$\begin{pmatrix} 0 & 0 & \dots & 0\\ \tau_1 & \tau_2 & \dots & \tau_{m_j}\\ 0 & 0 & \dots & 0\\ u(\tau_1) & u(\tau_2) & \dots & u(\tau_{m_j})\\ -\textbf{e}_j & -\textbf{e}_j & \dots & -\textbf{e}_j \end{pmatrix}$} \\
\multicolumn{1}{c|}{key} &time & & \\
\multicolumn{1}{c|}{} &space & & \\
\cline{1-2}
\multicolumn{2}{c|}{value} & & \\
\cline{1-2}
\multicolumn{2}{c|}{index} & & \\
\hline
\end{tabular}
  \hfill
    \caption{The values of condition and QoI in the $j$-th demo in the example of solving the one-dimensional forward ODE problem. Here the condition consists of $c\colon [0,T]\to \R$ and the initial condition $u(0)$; and $u \colon [0,T]\to \R$ is the QoI. We use $n_j-1$ key-value pairs to represent $c$, one key-value pair for $u(0)$,  and $m_j$ key-value pairs for $u$. Note that in the first row, we use the indicator $0$ and $1$ to distinguish different terms in the condition, i.e., $c$ and $u(0)$. The third row is populated with zeros since there are no spatial coordinates. $\textbf{e}_j$ is the column index vector.
}
    \label{tab:sec2_demok_linearode}
\end{table}

In the end, we remark that the number of demos and key-value pairs may differ across various prompts. Transformers are specifically designed to handle inputs of different lengths. However, for the purpose of batching, we still use zero-padding to ensure consistent lengths. Such padding, along with appropriate masks, has no impact on mathematical calculations.

\begin{figure}[!ht]
\centering
\includegraphics[width=0.9\textwidth]{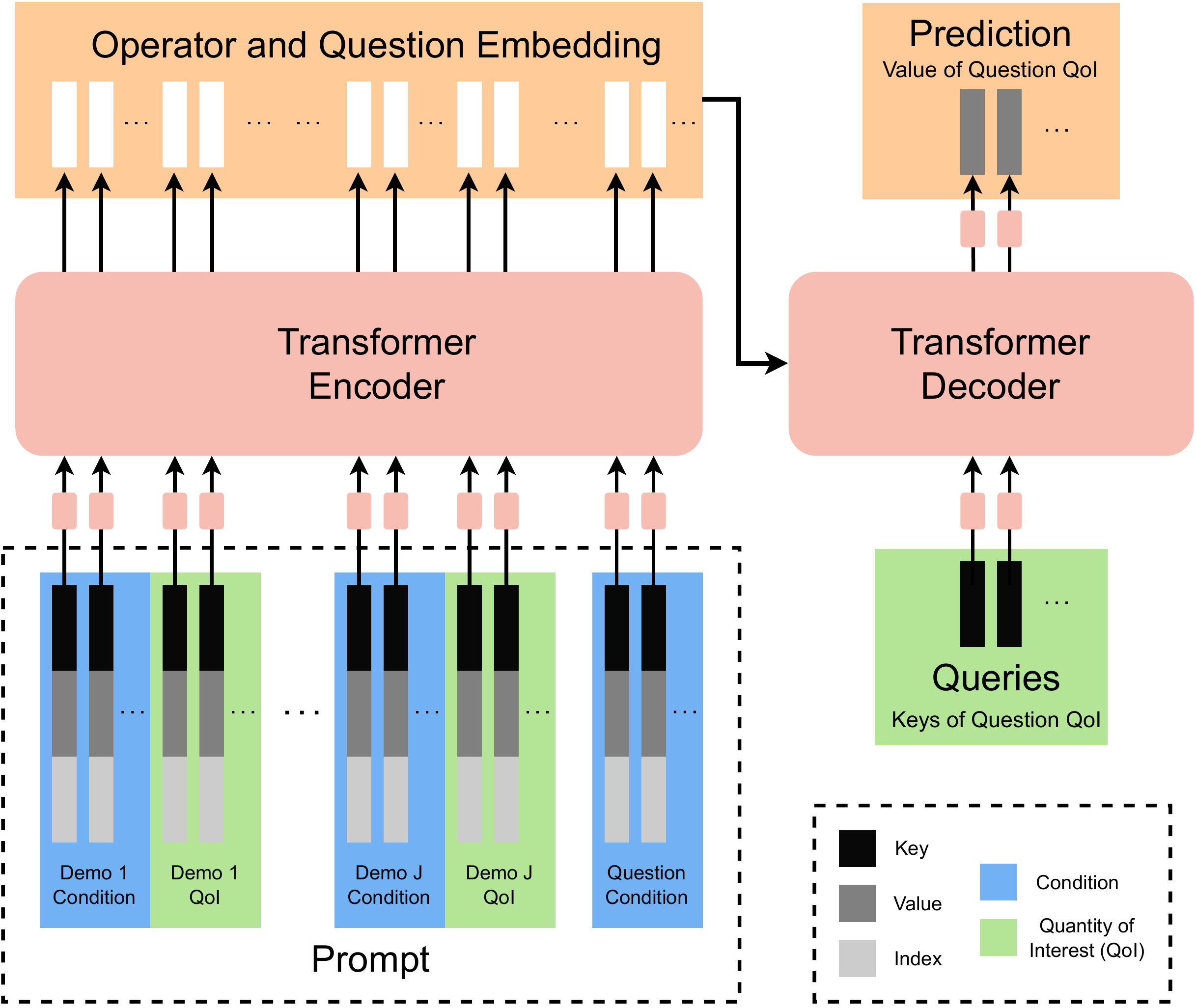}
\caption{The neural network architecture for In-Context Operator Networks (ICON).}
\label{fig:nn_architecture}
\end{figure}

\subsection{Neural Network Architecture}\label{sec:nn-architecture}
We employ an encoder-decoder neural network architecture in our method. The architecture of neural networks is shown in Figure~\ref{fig:nn_architecture}.

Before entering the encoder, the columns of the prompt are processed by a shared linear layer. The encoder, which is a self-attention transformer, combines information from all demos and the question condition within the prompt. This process generates an output that represents an embedding of the operator and the question. Along with the queries (which also undergo another shared linear layer), this embedding is fed into the decoder, which is a cross-attention transformer. Finally, the decoder's output is passed through an additional linear layer to adjust its dimensionality to match the question's quantity of interest~(QoI).

The encoder-decoder structure utilized in this architecture shares similarities with the one used in computer vision for object detection tasks~\cite{carion2020end}. In that particular context, the decoder takes the image embedding and ``object queries'' as inputs. Each output from the decoder is then forwarded to a common feed-forward network that predicts the detection.

Note that in our decoder, there are no self-attention layers for the queries. Therefore, with a fixed prompt, if we input $n$ query vectors (or $n$ keys of question QoI) into the model and receive $n$ corresponding values as output, each value is exclusively determined by its corresponding query, unaffected by the others. Such independence enables us to design an arbitrary number of queries, wherever we wish to evaluate the question QoI function.

\begin{algorithm}[htbp]
\SetAlgoLined
 \For{each type of problem}{
    Randomly generate $M$ sets of parameters;\\
    // Each set of parameters defines an operator\\
    \For{each set of paramters}{
        Randomly generate $N$ pairs of conditions and QoIs\;
        // These $N$ pairs of conditions and QoIs share the same operator
    }
 }
 \caption{Data preparation. \label{alg:data_preparation}}
\end{algorithm}

\begin{algorithm}[htbp]
\SetAlgoLined
 // Training stage:

 \For{$i = 1, 2, \dots,$ training steps}{
    \For{$b = 1, 2, \dots,$ batch size}{
        Randomly select a type of problem and a set of parameters from dataset\;
        Randomly set the number of examples $J$, and the number of key-value pairs in each condition and QoI of the examples and question\;
        From $N$ pairs of conditions and QoIs, randomly select $J$ pairs as examples and one pair as the question\;
        Build a prompt matrix, query vectors, and the ground truth using the selected examples and question\;
    }
    Use the batched prompts, queries and labels to calculate the MSE loss and update the neural network parameters with gradients\;
 }
 // Inference stage:
 
 Given a new system with an unknown operator, collect examples and a question condition, and design the queries\;
 Construct the prompt using the examples and question condition\;
 Get the prediction of the question QoI using a forward pass of the neural network\;
 \caption{The training and inference of In-Context Operator Networks (ICON). \label{alg:icon}}
\end{algorithm}

\subsection{Data Preparation and Training}\label{sec:training}
Before training the neural network, we prepare data that contains the numerical solutions to different kinds of differential equation problems. The details of data generation are described in Algorithm~\ref{alg:data_preparation}.
 
In the training process, in each iteration, we randomly build a batch of prompts, queries, and labels (ground truths) from data. Note that different problems with different operators appear in the same batch. The loss function is the mean squared error (MSE) loss between the outputs of the neural network and the labels. The details of the training process are described in Algorithm~\ref{alg:icon}.

\subsection{Inference: Few-shot Learning without Weight Update}\label{sec:inference}

After the training, we use the trained neural network to make predictions of the question QoI based on a few demos that describes the operator, as well as the question condition. 

During one forward pass, the neural network finishes the following two tasks simultaneously: it learns the operator from the demos, and applies the learned operator to the question condition for predicting the question QoI. We emphasize that the neural network does not update its weights during such a forward pass. In other words, the trained neural network acts as a few-shot operator learner, and the training stage can be perceived as ``learning to learn operators''.

\section{Numerical Results}
\label{sec:experiments}

We designed 19 types of problems for training, each of which has 1000 sets of parameters, so $19\times 1000 = 19000$ operators in total. For each operator, we generate $100$ condition-QoI pairs. In other words, $M = 1000, N = 100$ in Algorithm~\ref{alg:data_preparation} and ~\ref{alg:icon}.

\subsection{Problems}

In this subsection, we list all 19 types of problems, as well as the setups for parameters and condition-QoI pairs in the data preparation stage (Algorithm~\ref{alg:data_preparation}) in Table~\ref{tab:problem_table}.  As for the implementation of the parameters, we present them in the appendix section.

During training, we randomly select 1-5 demos when building the prompt. The number of key-value pairs in each condition/QoI randomly ranges from 41 to 50. The details are in the appendix.

\begin{landscape}
\begin{table}[!htbp]
\centering
\caption{List of differential equation problems.}
\label{tab:problem_table}

  \begin{tabular}{|l|l|p{6.6cm}|l|l|p{2.7cm}|}
    \hline
\# & Problem Description & \text{Differential Equations} & Parameters & Conditions & QoIs \\
\hline
1 & Forward problem of ODE 1 & \multirow{2}{*}{$\frac{d}{dt}u(t) = a_1 c(t) + a_2$ for $t \in[0,1]$} & \multirow{2}{*}{$a_1, a_2$} & $u(0), c(t), t\in[0,1]$ & $u(t), t\in[0,1]$ \\
\cline{1-2}\cline{5-6}
2 & Inverse problem of ODE 1 & & & $u(t), t\in[0,1]$ & $c(t), t\in[0,1]$ \\
\hline
3 & Forward problem of ODE 2 & \multirow{2}{*}{$\frac{d}{dt}u(t) = a_1 c(t) u(t) + a_2$ for $t \in[0,1]$} & \multirow{2}{*}{$a_1, a_2$} & $u(0),c(t), t\in[0,1]$ & $u(t), t\in[0,1]$ \\
\cline{1-2}\cline{5-6}
4 & Inverse problem of ODE 2 & & & $u(t), t\in[0,1]$ & $c(t), t\in[0,1]$ \\
\hline
5 & Forward problem of ODE 3 & \multirow{2}{*}{$\frac{d}{dt}u(t) = a_1 u(t) + a_2(t) c(t) + a_3$ for $t \in[0,1]$} & \multirow{2}{*}{$a_1, a_2, a_3$} & $u(0),c(t), t\in[0,1]$ & $u(t), t\in[0,1]$ \\
\cline{1-2}\cline{5-6}
6 & Inverse problem of ODE 3 & & & $u(t), t\in[0,1]$ & $c(t), t\in[0,1]$ \\
\hline
7 & Forward damped oscillator  & \multirow{2}{*}{$u(t) = A \sin(\frac{2\pi}{T} t + \eta)e^{-kt}$ for $t \in[0,1]$} & \multirow{2}{*}{$k$} & $u(t), t \in[0,0.5)$ & $u(t), t \in[0.5 , 1]$ \\
\cline{1-2}\cline{5-6}
8 & Inverse damped oscillator  & & & $u(t), t\in[0.5 , 1]$ & $u(t), t\in[0 , 0.5)$ \\
\hline
9 & Forward Poisson equation & \multirow{2}{*}{$\frac{d^2}{dx^2}u(x) = c(x)$ for $x\in[0,1]$} & \multirow{2}{*}{$u(0), u(1)$} & $c(x), x\in[0,1]$ & $u(x), x\in[0,1]$ \\
\cline{1-2}\cline{5-6}
10 & Inverse Poisson equation & & & $u(x), x\in[0,1]$ & $c(x), x\in[0,1]$ \\
\hline
11 & Forward linear reaction-diffusion & \multirow{2}{*}{{\parbox{6cm}{$-\lambda a\frac{d^2}{dx^2}u(x) + k(x)u(x) = c$ \\ for $x\in[0,1]$, $\lambda = 0.05$}}} & \multirow{2}{*}{$u(0), u(1), a,c$} & $k(x), x\in[0,1]$ & $u(x), x\in[0,1]$ \\
\cline{1-2}\cline{5-6}
12 & Inverse linear reaction-diffusion & & & $u(x), x\in[0,1]$ & $k(x), x\in[0,1]$ \\
\hline
13 & Forward nonlinear reaction-diffusion  & \multirow{2}{*}{{\parbox{6cm}{$- \lambda a \frac{d^2}{dx^2}u(x) + ku^3 = c(x)$ \\for $x\in[0,1]$, $\lambda = 0.1$}}} & \multirow{2}{*}{$u(0), u(1), k, a$} & $c(x), x\in[0,1]$ & $u(x), x\in[0,1]$ \\
\cline{1-2}\cline{5-6}
14 & Inverse nonlinear reaction-diffusion & & & $u(x), x\in[0,1]$ & $c(x), x\in[0,1]$ \\
\hline
 15 &   MFC $g$-parameter $1$D $\rightarrow 1$D & \multirow{5}{*}{\parbox{6cm}
{$\quad$\\
$\quad$\\
$\inf_{\rho, m} \iint c\dfrac{m^2}{2 \rho} dx dt + \int g(x) \rho(1,x) dx$\\
such that\\
$\partial_t \rho(t,x) + \nabla_x m(t,x) = \mu \Delta_x \rho(t,x)$
\\
for $t \in [0,1], x \in [0,1]$,\\
$c = 20, \mu = 0.02$,
periodic boundary condition in spatial domain
 
 }} & \multirow{3}{*}{$g(x), x\in[0,1]$} & $\rho(t=0,x), x\in[0,1]$ & {\parbox{2.7cm}{$\rho(t=1,x),\\ x\in[0,1]$}} \\
    \cline{1-2}\cline{5-6}
16 &     MFC $g$-parameter $1$D $\rightarrow 2$D  &  &  & $\rho(t=0,x), x\in[0,1]$ &{\parbox{2.7cm}{$\rho(t,x)$, $x\in[0,1]$,\\$t\in[0.5,1]$}} \\
    \cline{1-2}\cline{5-6}
 17 &    MFC $g$-parameter $2$D $\rightarrow 2$D  &  &  & {\parbox{2.8cm}{$\rho(t,x), t\in[0,0.5)$, \\ $x\in [0,1]$}} & \parbox{2.7cm}{$\rho(t,x)$, $x\in[0,1]$,  $t\in[0.5,1]$} \\
    \cline{1-2}\cline{4-6}
18 &   MFC $\rho_0$-parameter $1$D $\rightarrow 1$D  &  & \multirow{2}{*}{{\parbox{2.7cm}{$\rho(t=0,x),$\\ $x\in[0,1]$}}} & \multirow{2}{*}{$g(x), x\in[0,1]$} & {\parbox{2.7cm}{$\rho(t=1,x)$, \\ $x\in[0,1]$} }\\
    \cline{1-2}\cline{6-6}
 19 &  MFC $\rho_0$-parameter $1$D $\rightarrow 2$D  &  &  &  & {\parbox{2.7cm}{$\rho(t,x)$, $x\in[0,1],\\t\in[0.5,1]$ }}\\
\hline
\end{tabular}
\end{table}
\end{landscape}

\begin{figure}[!ht]
\centering
\begin{subfigure}[b]{0.32\textwidth}
\centering
\includegraphics[width=\textwidth]{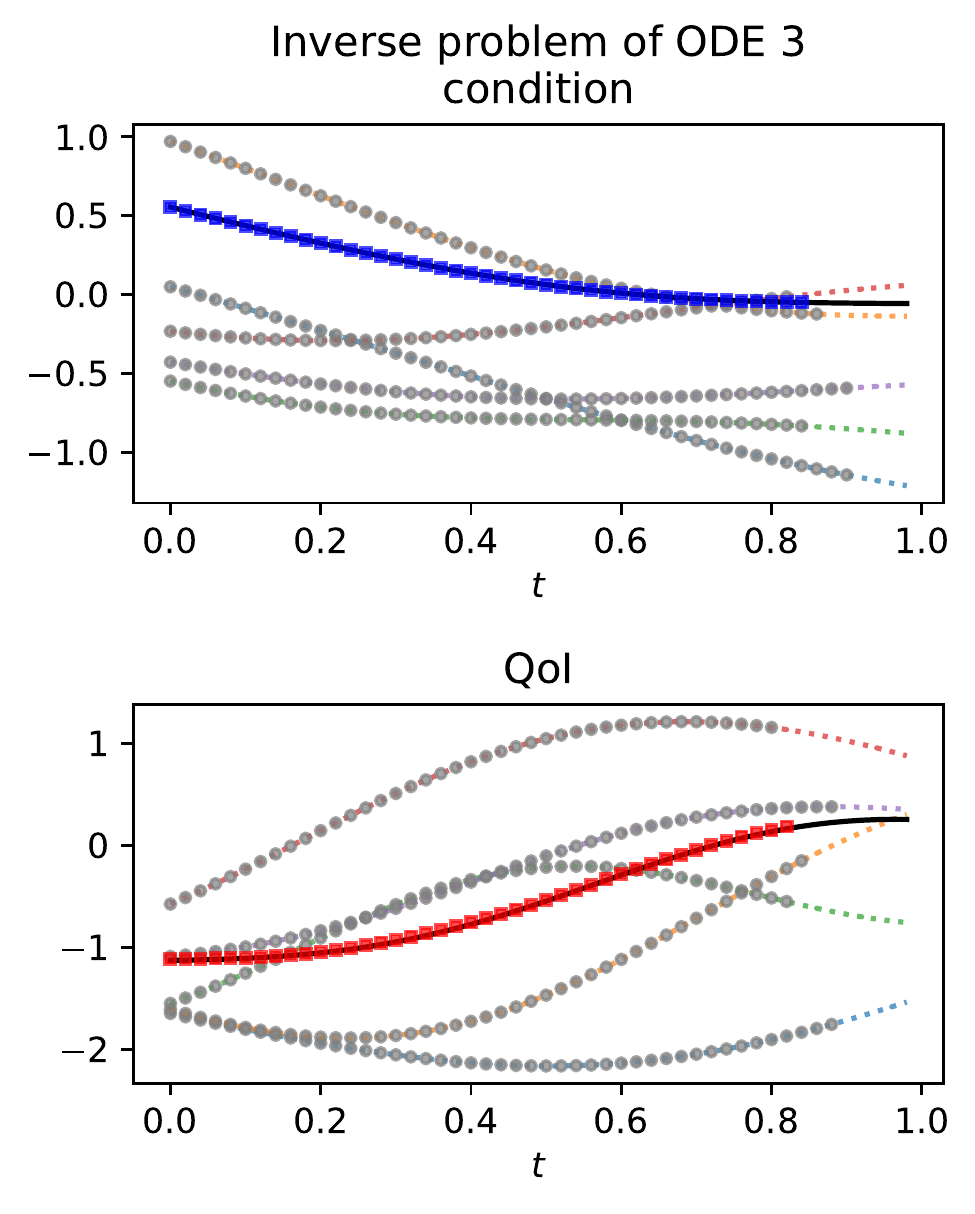}
\caption{}
\label{fig:ind_subfigure1}
\end{subfigure}
\begin{subfigure}[b]{0.32\textwidth}
\centering
\includegraphics[width=\textwidth]{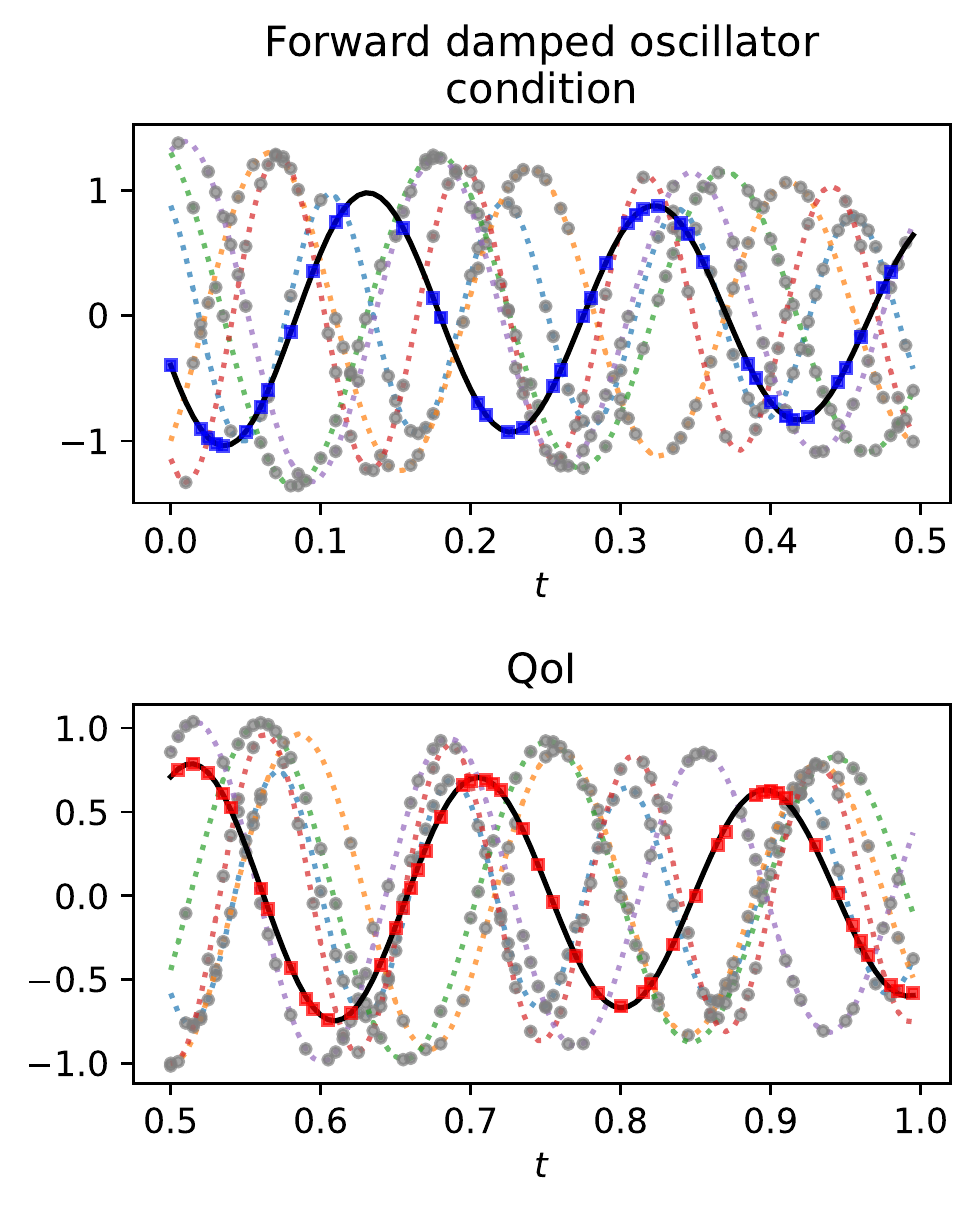}
\caption{}
\label{fig:ind_subfigure2}
\end{subfigure}
\begin{subfigure}[b]{0.32\textwidth}
\centering
\includegraphics[width=\textwidth]{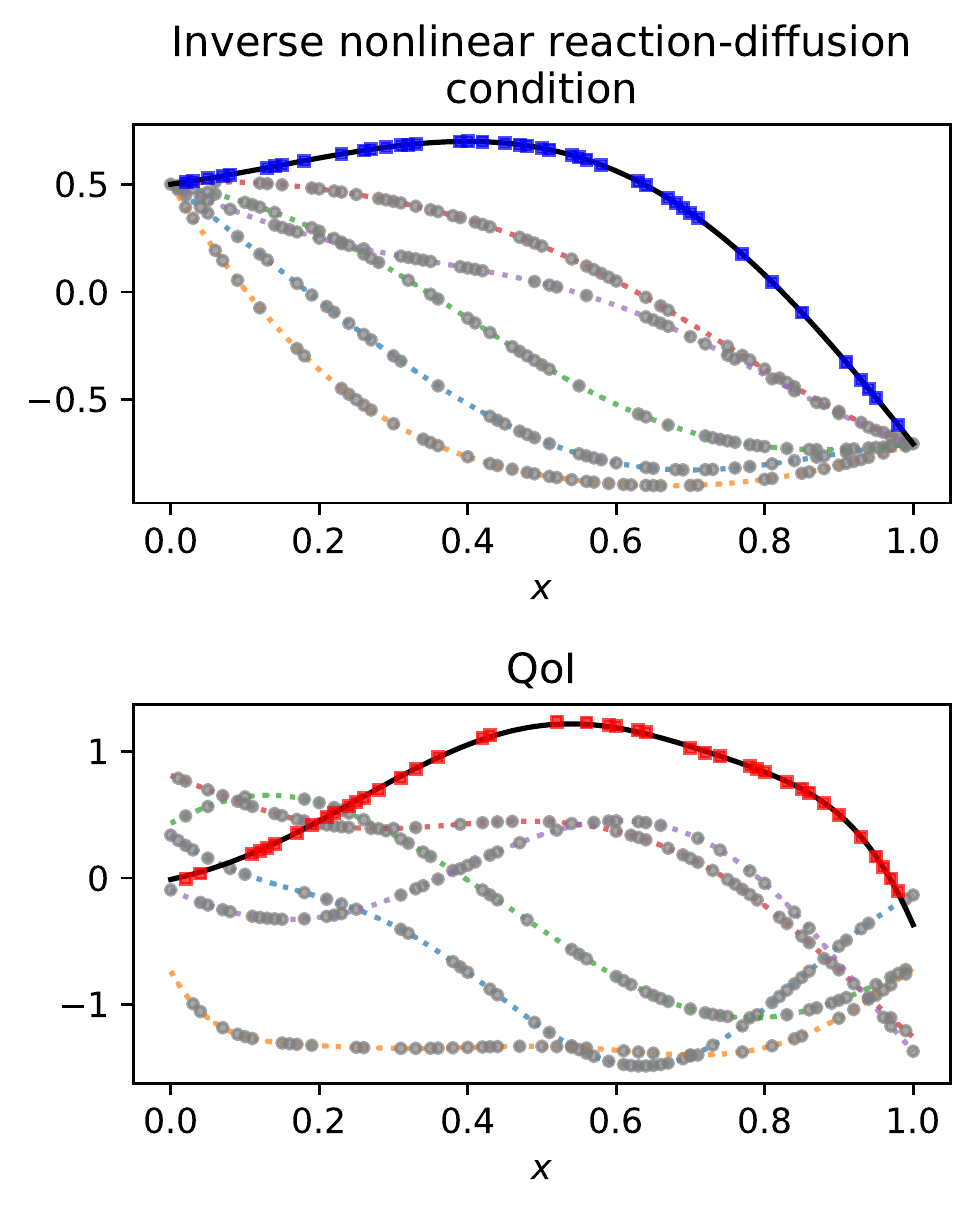}
\caption{}
\label{fig:ind_subfigure3}
\end{subfigure}
\caption{
Visualization of three in-context operator learning test cases for the selected differential equation problems. The problem type is shown in the title. The colored dotted line represents the hidden function of conditions and QoIs in demos, while the grey dots represent the sampled key-value pairs of the demo conditions and QoIs used in the prompts. The blue dots represent the key-value pairs in the question conditions, sampled from the hidden function of the question condition in black solid lines. The neural network prediction of the question QoI is illustrated with red dots. One can see the consistency between the prediction and the ground truth (solid black lines). 
}
\label{fig:visualize_1D}
\end{figure}

\begin{figure}[!ht]
\centering
\includegraphics[width=0.32\textwidth]{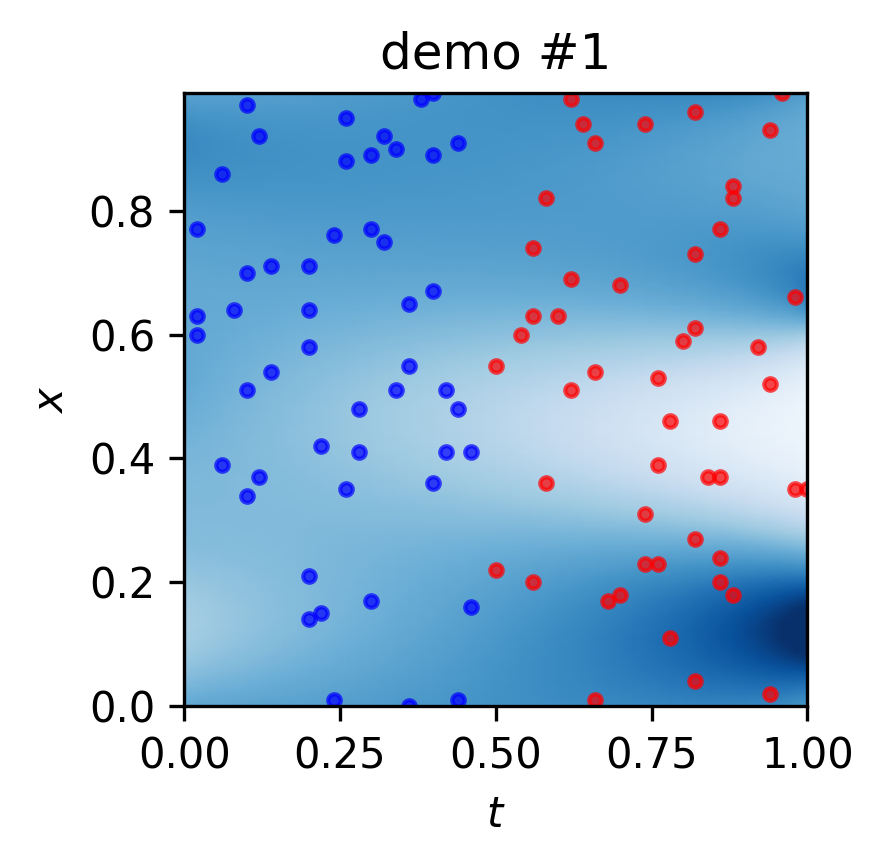}
\includegraphics[width=0.32\textwidth]{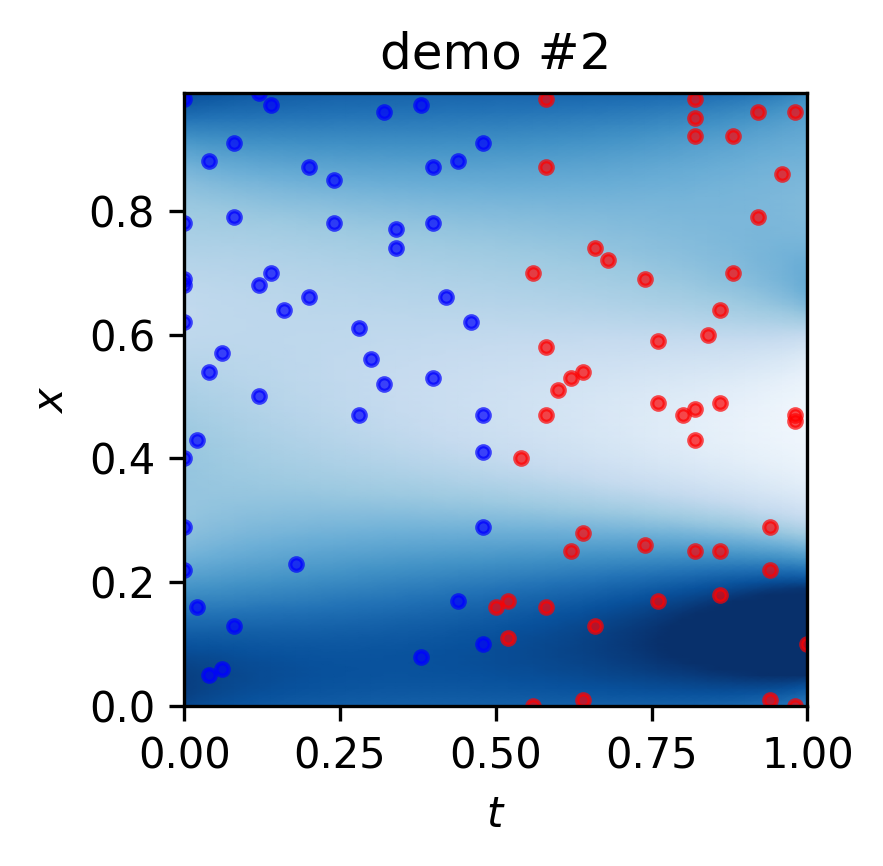}
\includegraphics[width=0.32\textwidth]{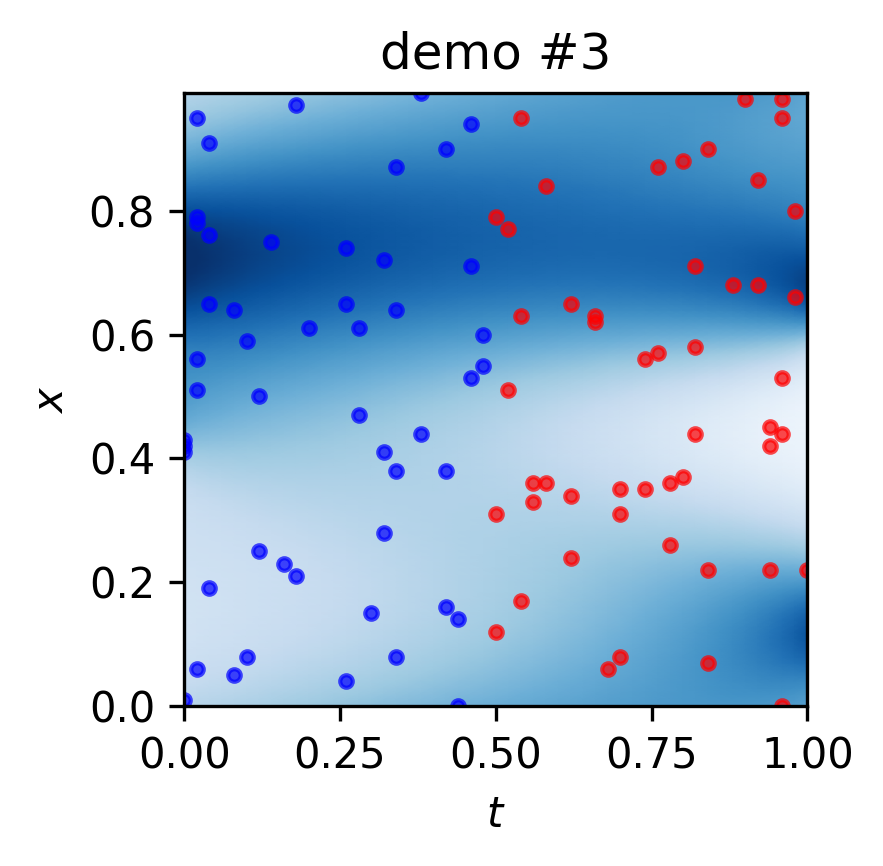}
\includegraphics[width=0.99\textwidth]{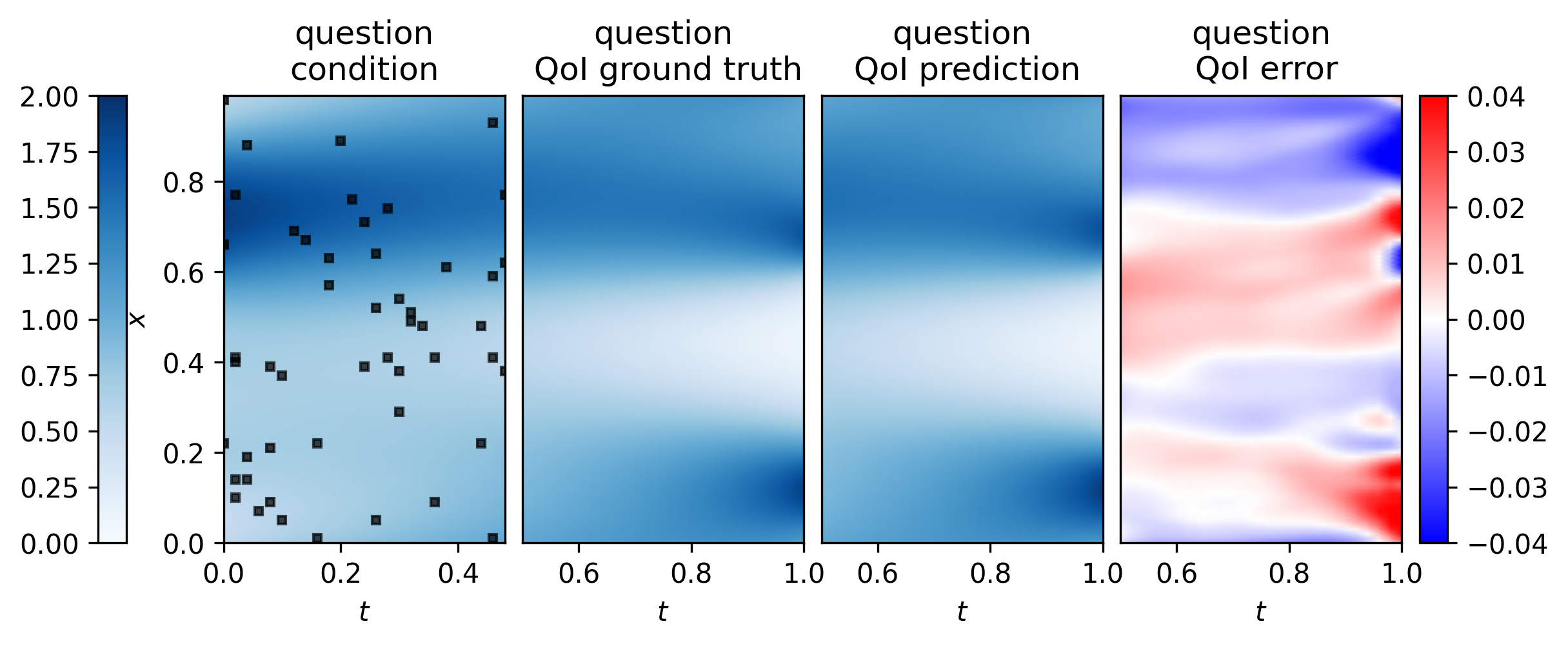}
\caption{Visualization of the in-context operator learning test case for problem 17: mean-field control (MFC) $g$-parameter $2$D $\rightarrow 2$D. On the top, we show the three demos used when building the prompt, where the blue dots represent the sampled key-value pairs of conditions; the red dot represents the sampled key-value pairs of QoIs. At the bottom, we present the question condition (black dot indicates the condition), ground truth, prediction, and the errors (difference between prediction and ground truth). Note here, we obtain the prediction of the density profile for $t\in [0.5,1], x\in [0,1]$ by setting the question keys as grid points over the temporal-spatial domain. The demos and question conditions/QoIs share the same color bar.}
\label{fig:visualize_2D}
\end{figure}

\begin{figure}[!ht]
\centering
\includegraphics[width=0.32\textwidth]{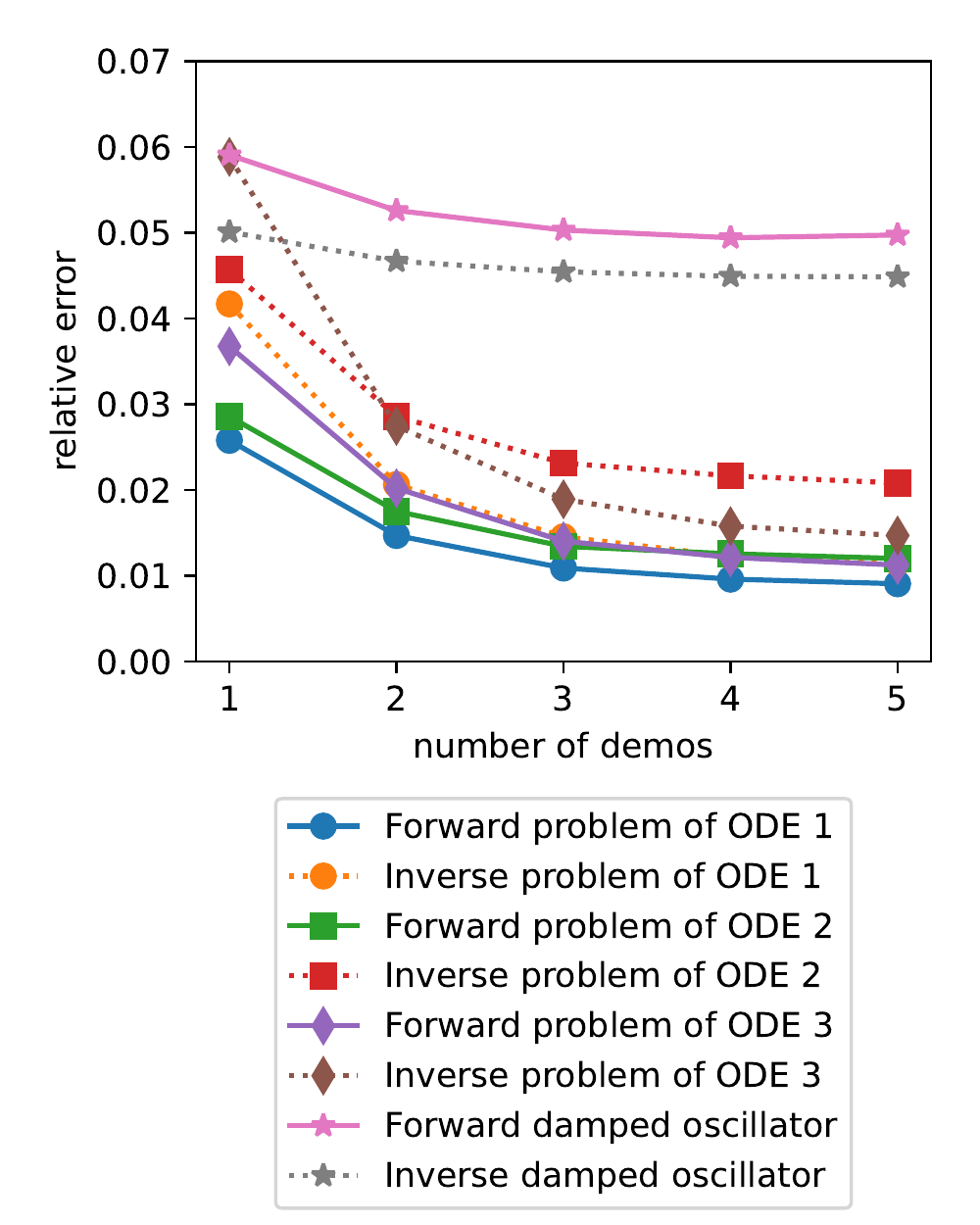}
\includegraphics[width=0.32\textwidth]{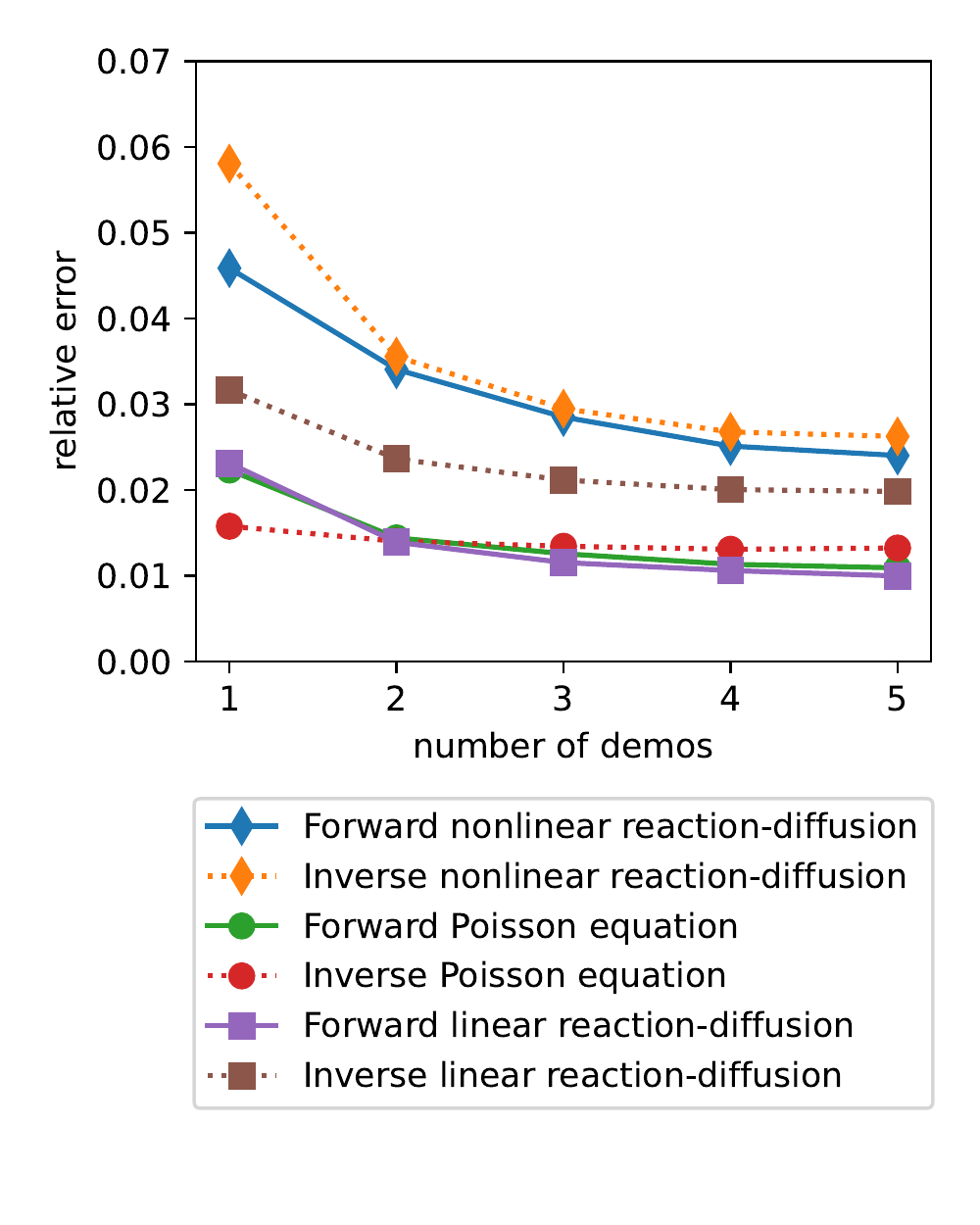}
\includegraphics[width=0.32\textwidth]{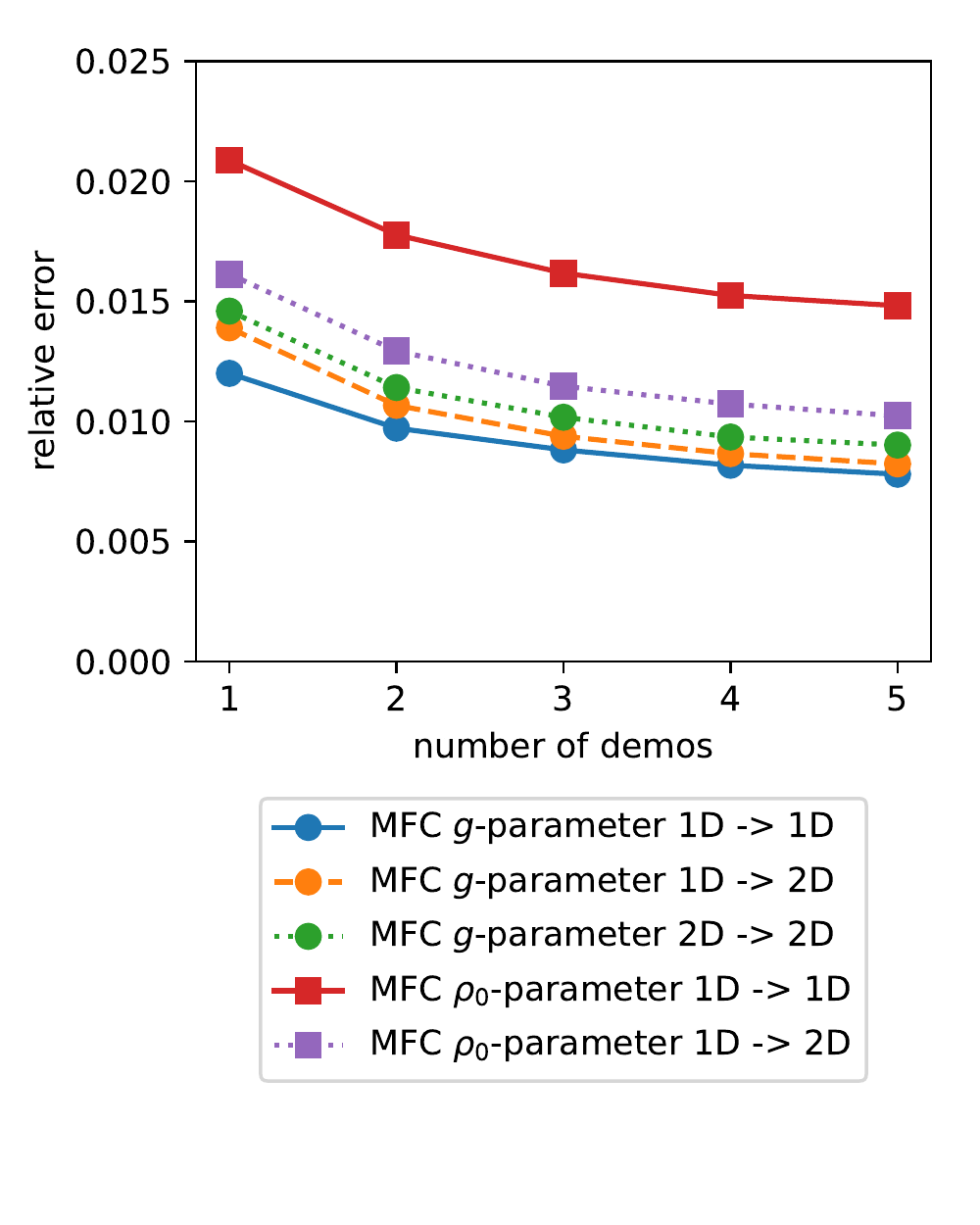}
\caption{Average relative in-distribution testing errors for all problems listed in Table~\ref{tab:problem_table}. The error decreases with the number of demos in each prompt.}
\label{fig:results_err_ind}
\end{figure}

\subsection{In-Distribution Operators}\label{sec:ind_error}

In this section, we show the testing errors for each of the 19 types of problems, with the distributions for parameters, conditions and QoIs the same as in the training stage, i.e., in-distribution operator learning. The number of key-value pairs in each condition/QoI randomly ranges from 41 to 50, as in the training stage. By using different random seeds, we ensure that the testing data are different from the training data (although in the same distribution), and that each condition-QoI pair only shows once, either as a demo or a question, during testing. 

We show several in-context operator learning test cases in Figure~\ref{fig:visualize_1D} and Figure~\ref{fig:visualize_2D}.

In Figure~\ref{fig:results_err_ind}, we show relative errors with respect to the number of demos in each prompt for all $19$ problems listed in Table~\ref{tab:problem_table}. For each type of problem, we conduct $500$ in-context learning cases, corresponding to $100$ different operators, i.e., $5$ cases for each operator. Firstly, the absolute error is computed by averaging the differences between the predicted question QoI values and their corresponding ground truth values across all in-context learning cases. Then, the relative error is obtained by dividing the absolute error by the mean of the absolute values of the ground truth values.

Across all 19 problems examined, it is evident from Figure~\ref{fig:results_err_ind} that the average relative error remains below 6\% even in the cases with a single demo. The majority of the average relative falls around 2\% when using 5 demonstrations. This underscores the capacity of a single neural network to effectively learn the operator from demos and accurately predict the QoI for various types of differential equation problems.
Furthermore, the error consistently decreases as the number of demos in each prompt increases for all 19 problems.

\subsection{Super-Resolution and Sub-Resolution}

\begin{figure}[!ht]
\centering
\includegraphics[width=0.6\textwidth]{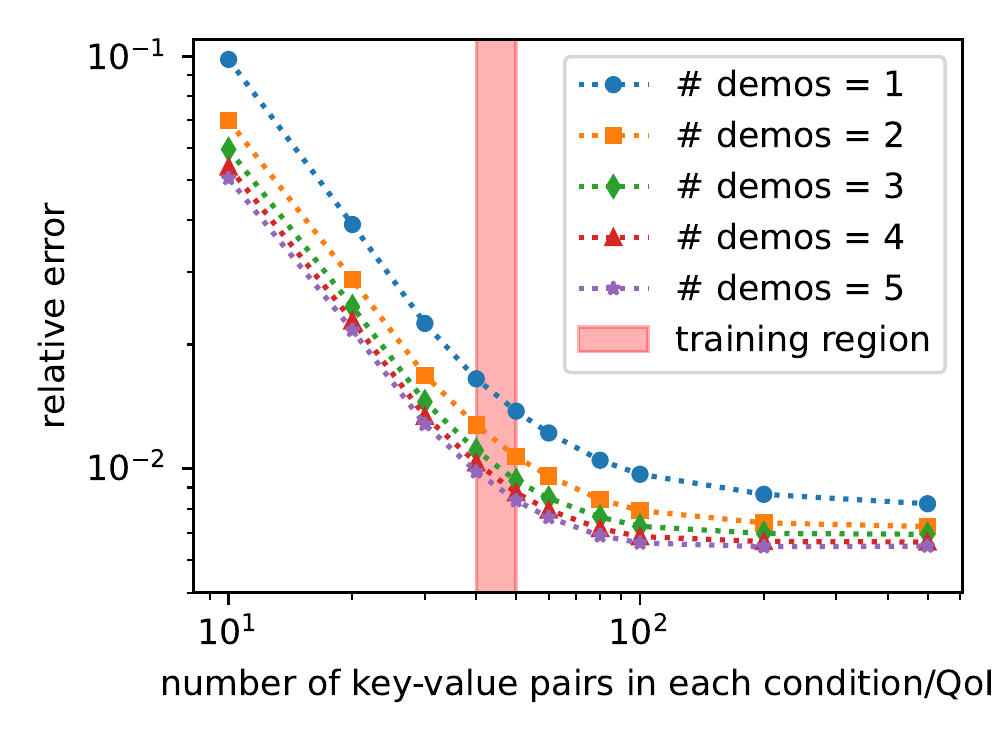}
\caption{Average relative testing errors for problem 17, i.e., MFC $g$-parameter $2$D $\rightarrow 2$D, with the number of key-value pairs ranging from 10 to 500 in each condition/QoI. As we increase the number of key-value pairs, the error decreases and finally converges below $1\%$. Note that the neural network is trained using 41 to 50 key-value pairs, represented by the narrow red region in the figure.}
\label{fig:results_err_length}
\end{figure}

Even though the neural network is trained using 41 to 50 key-value pairs to represent conditions and QoIs, it demonstrates the ability to generalize to a significantly broader range of numbers without requiring any fine-tuning, including more key-value pairs (super-resolution) or less key-value pairs (sub-resolution).

In figure~\ref{fig:results_err_length}, we examine the neural network on problem 17, i.e. mean-field control (MFC) $g$-parameter $2$D $\rightarrow 2$D, with the number of (randomly sampled) key-value pairs ranging from 10 to 500 in each condition/QoI. The average relative error is calculated in the same way as in Section~\ref{sec:ind_error}, except that we make predictions and evaluate errors in the domain $(t,x) \in [0.5,1]\times[0,1]$, by setting the question
keys as grid points over the temporal-spatial domain. A case of 3 demos, and 50 key-value pairs is illustrated in Figure~\ref{fig:visualize_2D}.

With a fixed number of demos in the prompt, the average relative error decreases with an increasing number of key-value pairs in each condition/QoIs, and finally converges below 1\%, even for the case of a single demo.

\subsection{Out-of-Distribution Operators}\label{sec:ood}

In this section, we examine the capability of the neural network in generalizing in-context learning to operators beyond the training distribution. Here we emphasize that the term ``out-of-distribution'' does not refer to the conditions, but rather to the operator itself being outside the distribution of operators observed during training.

\begin{figure}[ht]
\centering
\begin{subfigure}[b]{0.48\textwidth}
\centering
\includegraphics[width=\textwidth]{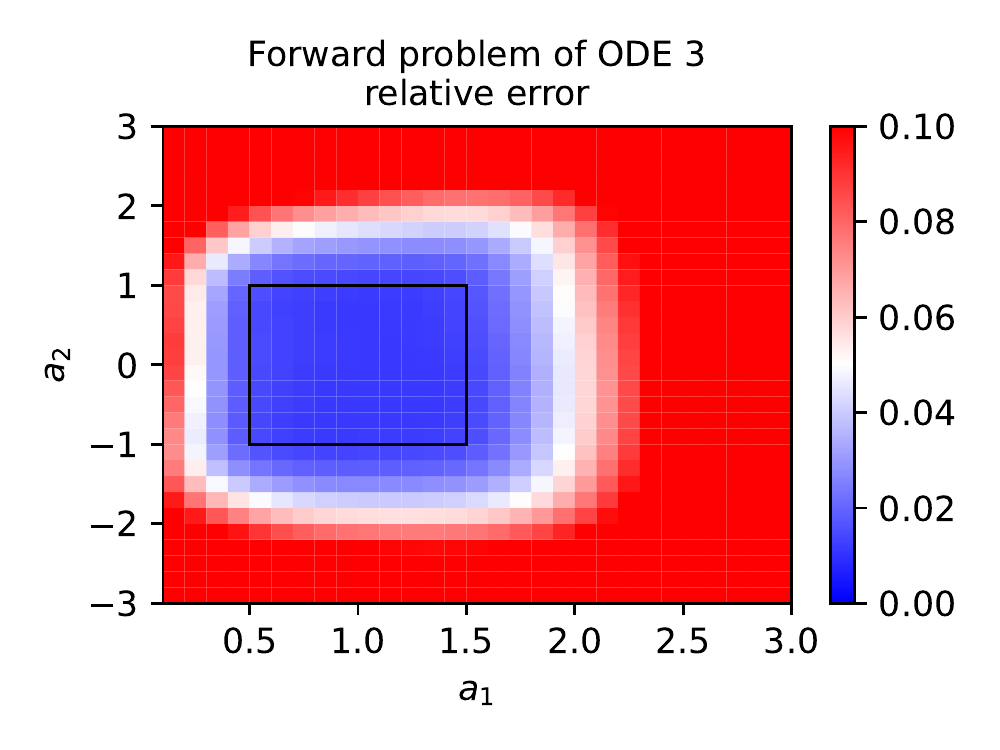}
\caption{}
\label{fig:results_err_ood_a}
\end{subfigure}
\begin{subfigure}[b]{0.48\textwidth}
\centering
\includegraphics[width=\textwidth]{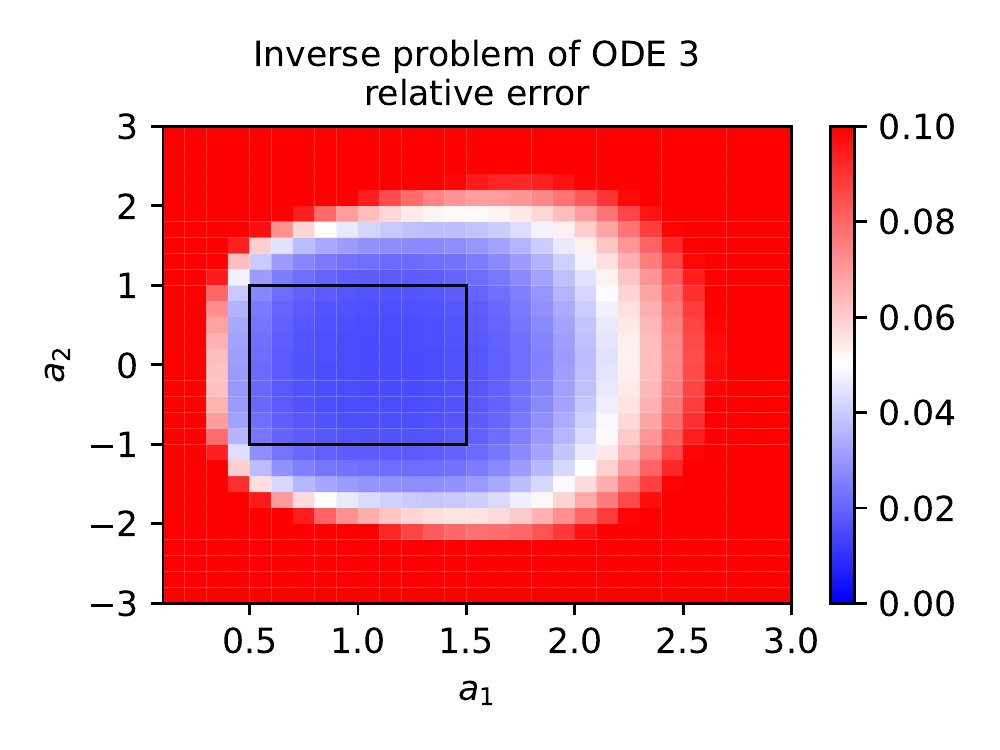}
\caption{}
\label{fig:results_err_ood_b}
\end{subfigure}
\begin{subfigure}[b]{0.48\textwidth}
\centering
\includegraphics[width=\textwidth]{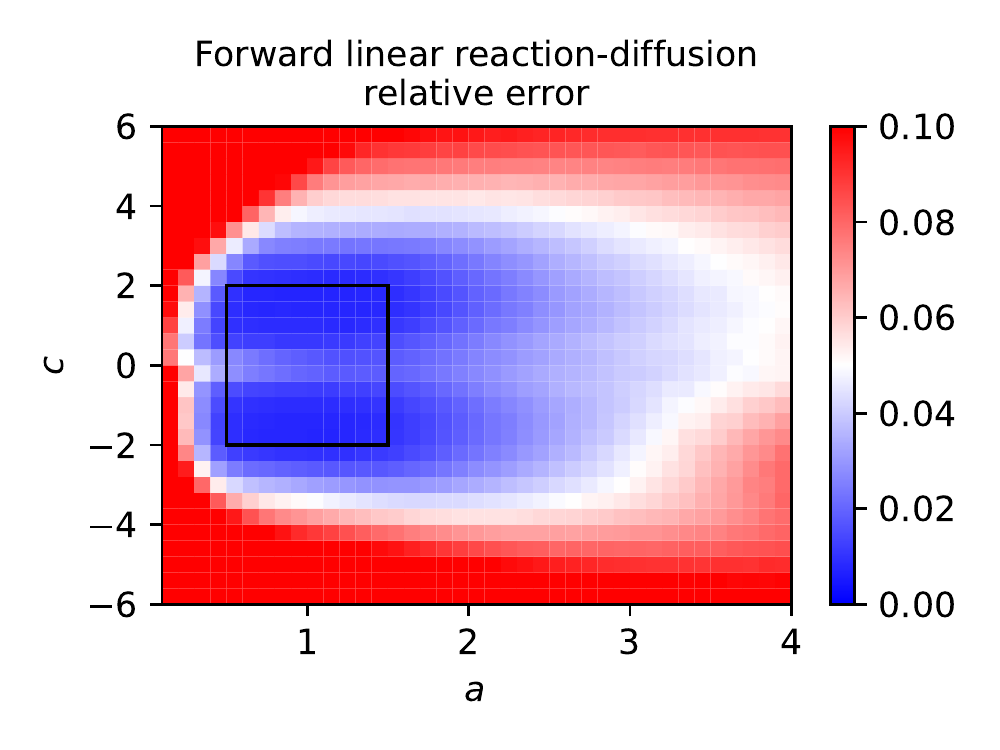}
\caption{}
\label{fig:results_err_ood_c}
\end{subfigure}
\begin{subfigure}[b]{0.48\textwidth}
\centering
\includegraphics[width=\textwidth]{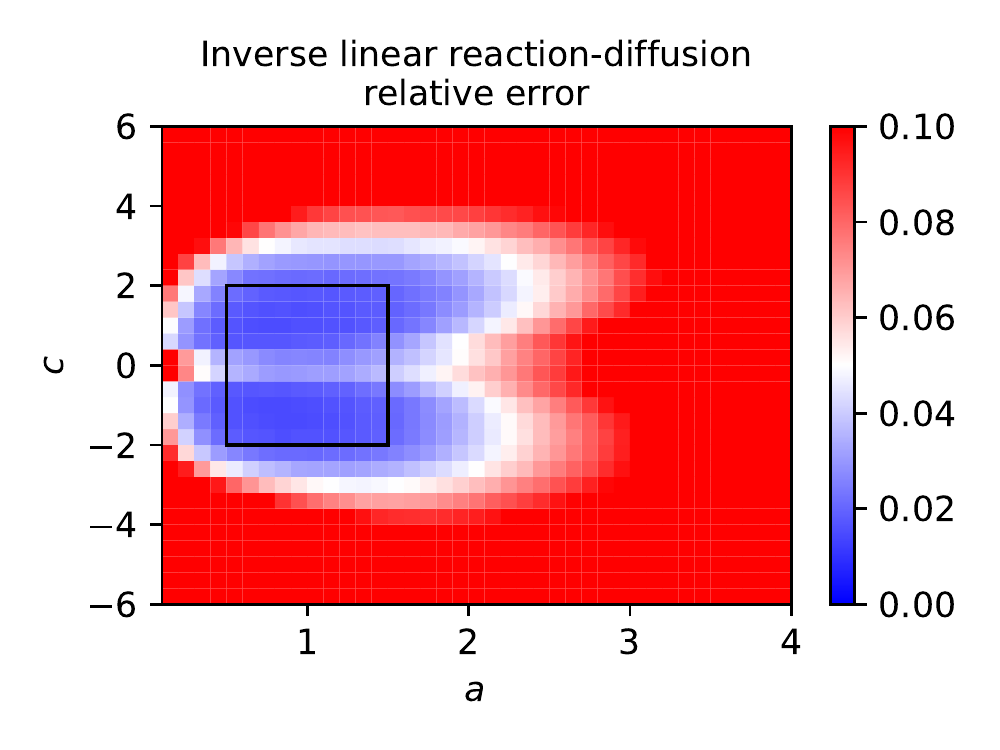}
\caption{}
\label{fig:results_err_ood_d}
\end{subfigure}
\caption{Average relative error for out-of-distribution operators. The region of operator parameters utilized during training is indicated by a black rectangle.}
\label{fig:results_err_ood}
\end{figure}

We conducted tests on four representative problem types, i.e. problems 5, 6, 11, and 12 in Table~\ref{tab:problem_table}. During the training process of the forward and inverse problems of ODE 3, we randomly generated $a_1$ and $a_2$ from a uniform distribution $\mathcal{U}(0.5,1.5)$ and $\mathcal{U}(-1,1)$, respectively. Each pair of $(a_1,a_2)$ defines an operator. Now, we expand the distribution of $(a_1,a_2)$ to a much larger region. To evaluate the performance, we divided the region $[0.1,3.0]\times[-3,3]$ into a grid and tested the performance within each cell. Specifically, we conducted $500$ in-context learning cases in each cell, corresponding to $100$ different operators and $5$ cases with different demos and questions for each operator. Here, the number of demos is fixed as five, and the number of key-value pairs is fixed as the maximum number used in training. We calculate the relative error for each cell and depict the results in Figure~\ref{fig:results_err_ood_a} and~\ref{fig:results_err_ood_b}. 

A similar analysis was applied to the forward and inverse problems of linear reaction-diffusion PDE problems. We divided the region of $(a,c)$ into a grid, while keeping the boundary condition parameters $u(0)$ and $u(1)$ randomly sampled from $\mathcal{U}(-1,1)$. The average relative errors are shown in Figure~\ref{fig:results_err_ood_c} and~\ref{fig:results_err_ood_d}. 

It is evident that for all four problems, the neural network demonstrated accurate prediction capabilities even with operator parameters extending beyond the training region. This showcases its strong generalization ability to learn and apply out-of-distribution operators.

\subsection{Generalization to Equations of New Forms}
As discussed in~\cite{brown2020language}, one of the advantages of in-context learning over pre-training plus fine-tuning is the ability to mix together multiple skills to solve new tasks. GPT-4~\cite{openai2023gpt4} even showed emergent abilities or behaviors beyond human expectations.

Although the scale of our experiment is much smaller than GPT-3 or GPT-4, we also observed preliminary evidence of the neural network's ability to learn and apply operators for equations of new forms that were never seen in training data.

\begin{figure}[ht]
\centering
\includegraphics[width=0.47\textwidth]{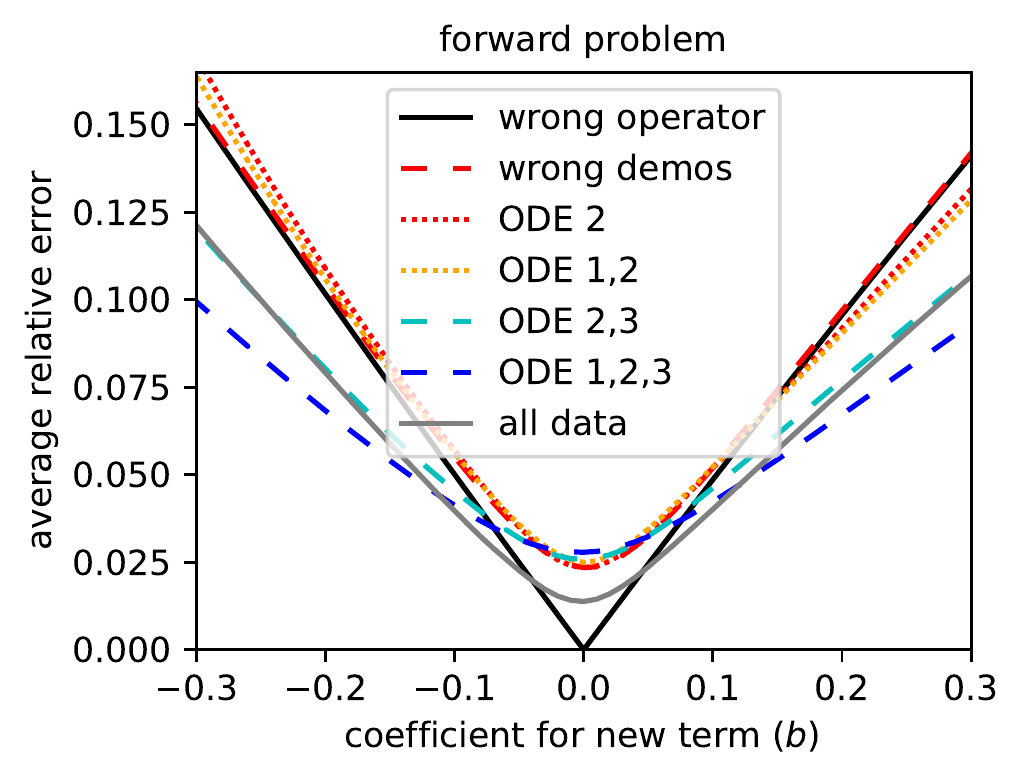}
\includegraphics[width=0.45\textwidth]{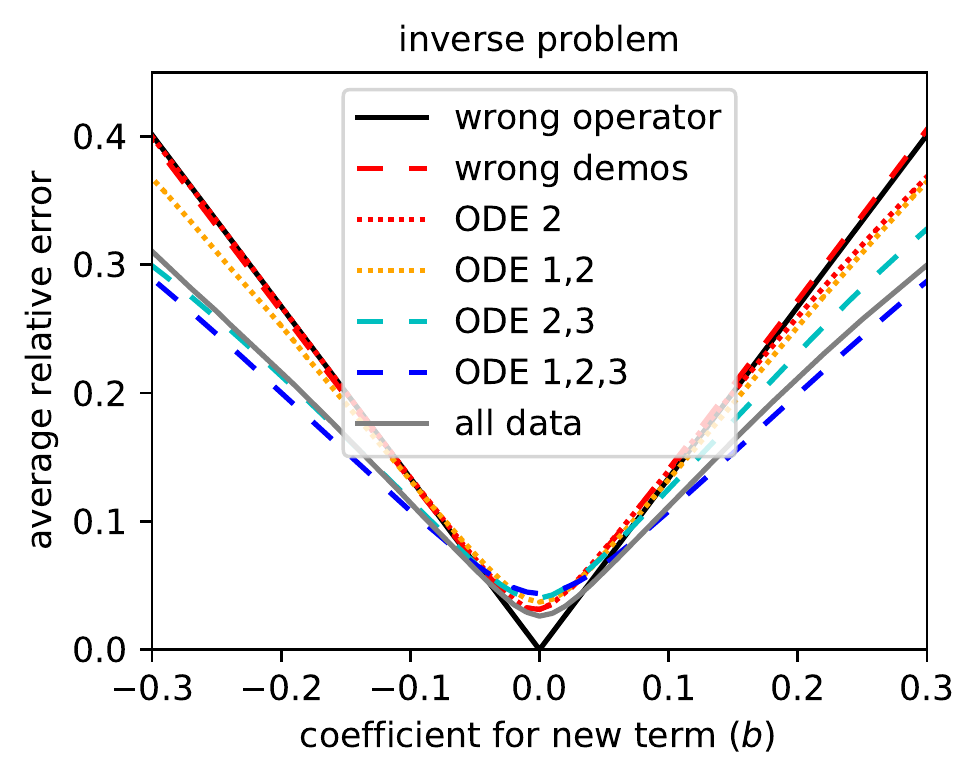}
\caption{Average relative errors for the new ODE, with the same neural network trained with different datasets. The error shows a decreasing trend as the training dataset becomes larger.}
\label{fig:results_err_nt}
\end{figure}

In particular, we designed a new ODE $u'(t) = a_1 u(t)c(t) + bu(t) + a_2$ over time interval $[0,1]$, by adding a linear term $bu(t)$ to ODE 2, which is borrowed from ODE 3. In the new problem, $b$ is also a parameter, and the operator is determined by $(a_1, a_2, b)$. We study the forward and inverse problems for the new ODE and evaluate performance of the neural network with $b\in [-0.3,0.3]$. The other setups, including the distribution of $a_1, a_2$ and $c(t)$, are the same as in problems 3 and 4 (forward and inverse problem of ODE 2). 

To study the influence of scaling up the training dataset, in Figure~\ref{fig:results_err_ood}, we show the average relative errors of neural networks trained with different training datasets. Here we obtain the average relative error for each $b$ in the same way as we did for each cell in Section~\ref{sec:ood}. To reduce the computational cost, in this section, we train the same neural network as the one analyzed in other sections, but only with half batch size, for 1/5 training steps. We remark that in these runs, the training datasets have different sizes, but the training steps and batch size remain the same. In other words, the neural network encounters the same number of prompts during training. The expansion of dataset types simply enhances the diversity of prompts.

We first train the neural network only with the datasets involving ODE 2 (both forward and inverse problems). Then, as a reference, we apply the ``wrong'' operator directly to the question condition. The ``wrong'' operator is defined as the one corresponding to ODE 2 $u'(t) = a_1 u(t)c(t) + a_2$ instead of the new ODE, with the same $a_1$ and $a_2$. Note that when $b=0$, the new ODE is reduced to ODE 2, thus the error is zero. As another reference, we perform in-context operator learning with the same neural network, but replace the demos in the prompts with the ones corresponding to ODE 2, denoted as ``wrong demos''. We can see the neural network with ``correct demos'' performs no better than both references, indicating that the network can hardly generalize its capability of in-context operator learning beyond ODE 2.

We then gradually add more ODE-related datasets to the training data. It is encouraging to see that the error shows a decreasing trend as the training dataset becomes larger. When trained with all ODE 1, 2, and 3, the neural network performs significantly better than the one merely trained with ODE 2. Such evidence shows the potential of the neural network to learn and apply operators corresponding to previously unseen equation forms, as we scale up the size and diversity of related training data.

In the end, we also show the results of the neural network used in other sections, which is trained with the full dataset with a larger batch size for a longer time. The performance on the new ODE is not improved, which is reasonable, since the newly added data on the damped oscillator, PDEs, and MFC problems are not closely related to the new ODE.

\section{Discussion}\label{sec:discussion}

Why a very few demos are sufficient to learn the operator? We try to answer this question as follows:
\begin{enumerate}
\item We actually only need to learn the operator for a certain distribution of question conditions, not for all possible question conditions.
\item The training operators and testing operators share commonalities. For example, for ODE problems, $u$'s time derivative, $u$, and $c$ satisfy the same equation at each time $t$. If the neural network captures such shared property during training, and also notices this property in the demos during inference, it only needs to identify the ODE, for which a few demos are sufficient. 
\item The operators in this paper are rather simple and limited to a small family, hence easy to identify with a few demos. It is likely that for a larger family of operators in training and testing, in-context operator learning requires more demos (especially for those complicated operators), as well as a larger neural network with more computation resources.
\end{enumerate}

\section{Summary}\label{sec:summary}

In this paper, we proposed In-Context Operator Networks (ICON) to learn operators for differential equation problems. It goes beyond the conventional paradigm of approximating solutions for specific problems or some particular solution operators. Instead, ICON acts as an ``operator learner'' during inference, i.e., learns an operator from the given data and applies it to new conditions without any weight updates.

Through our numerical experiments, we demonstrated that a single neural network has the capability to learn an operator from a small number of prompted demos and effectively apply it to the question condition. Such a single neural network, without any retraining or fine-tuning, can handle a diverse set of differential equation problems, including forward and inverse problems of ODEs, PDEs, and mean field control problems. Moreover, while the numbers of key-value pairs for representing the condition/QoI functions are limited to a narrow range during training, ICON can generalize its in-context operator learning ability to a significantly broader range during testing, with errors decreasing and converging as we increase the number of key-value pairs. Furthermore, ICON showed its capacity to learn operators with parameters that extend beyond the training distribution. In the end, our observations provide preliminary evidence of ICON's potential to learn and apply operators corresponding to previously unseen equation forms.

The scale of our experiments is rather small. In the future, we wish to scale up the size of the neural network, the types of differential equation problems, the dimensions of keys and values, the length of conditions and QoIs, and the capacity of demo numbers.
This requires further development of in-context operator learning, including improvements in neural network architectures and training methods, as well as further theoretical and numerical studies of how in-context operator learning works.
In the field of NLP, for example in GPT-4, scaling up leads to emergent abilities or behaviors beyond human expectations~\cite{openai2023gpt4}. 
We anticipate the possibility of witnessing such emergence in a large-scale operator learning network.

\section*{Acknowledgement}
This work is partially funded by AFOSR MURI FA9550-18-502, ONR N00014-18-1-2527,
N00014-18-20-1-2093 and N00014-20-1-2787.

\section*{Appendix}

\subsection{Details on Differential Equation problems}

In Table~\ref{tab:problem_para} below, we provide comprehensive information regarding the operator, condition, and QoI parameters for differential equation problems.

\begin{table}[!htp]
  \centering
  \begin{tabular}{|l|p{6cm}|p{6cm}|}
    \hline
    \# & Operator Parameters & Condition and QoI Paramters\\
    \hline
    1-2 & $a_1 \sim \mathcal{U}(0.5,1.5)$, i.e., randomly sampled from the uniform distribution $[0.5, 1.5]$, $a_2 \sim \mathcal{U}(-1,1)$. & $u(0)\sim \mathcal{U}(-1,1)$, $c(t) \sim \mathcal{GP}(0, k(x,x'))$, i.e., sampled from a Gaussian process with zero mean and kernel $k(x,x')$, where $k(x,x') = \sigma^2 \exp\left(-\frac{1}{2l^2}(x-x')^2\right)$, $\sigma^2 = 1$, $l=0.5$. \\
    \hline
    3-4 & $a_1 \sim \mathcal{U}(0.5,1.5)$, $a_2 \sim \mathcal{U}(-1,1)$. & $u(0)\sim \mathcal{U}(-1,1)$, $c(t) \sim \mathcal{GP}(0, k(x,x'))$ \\
    \hline
    5-6 & $a_1 \sim \mathcal{U}(-1,1)$, $a_2 \sim \mathcal{U}(0.5,1.5)$, $a_3 \sim \mathcal{U}(-1,1)$. & $u(0)\sim \mathcal{U}(-1,1)$, $c(t) \sim \mathcal{GP}(0, k(x,x'))$ \\
    \hline
    7-8 & $k\sim \mathcal{U}(0,2)$ & $A\sim \mathcal{U}(0.5,1.5)$, $T\sim \mathcal{U}(0.1,0.2)$, $\eta \sim \mathcal{U}(0,2 \pi)$ \\
    \hline
    9-10 & $u(0) \sim \mathcal{U}(-1,1)$, $u(1)\sim \mathcal{U}(-1,1)$. & $c(x) \sim \mathcal{GP}(0, k(x,x'))$, where $\sigma^2 = 2$, $l=0.5$. \\
    \hline
    11-12 & $u(0) \sim \mathcal{U}(-1,1)$, $u(1)\sim \mathcal{U}(-1,1)$, $a \sim \mathcal{U}(0.5,1.5)$, $c \sim \mathcal{U}(-2,2)$ & $k(x) = \text{softplus}\left(\hat{k}(x)\right)$, where $\hat{k}(x) \sim \mathcal{GP}(0, k(x,x'))$, where $\sigma^2 = 1$, $l=0.5$. \\
    \hline
    13-14 & $u(0) \sim \mathcal{U}(-1,1)$, $u(1)\sim \mathcal{U}(-1,1)$, $a \sim \mathcal{U}(0.5,1.5)$, $k \sim \mathcal{U}(0.5, 1.5)$ & $\hat{u}(x) \sim \mathcal{GP}(0, k(x,x'))$, where $\sigma^2 = 1$, $l=0.5$. We apply an affine map to $\hat{u}(x)$ to obtain $u(x)$ such that it satisfies the boundary condition $u(0)$, $u(1)$. Then we apply $u$ to the equation to obtain the term $c(x)$. \\
    \hline
    15-17 & $\hat{g}(x) \sim \mathcal{GP}(0, k_p(x,x'))$, where the Gaussian process with zero mean and kernel $k_p(x,x') =$ { \fontsize{4}{14}\selectfont $\sigma^2 \exp\left(-\frac{\left( \sin\left(2 \pi x\right)- \sin\left(2 \pi x'\right)\right)^2 + \left( \cos\left(2 \pi x\right)- \cos\left(2 \pi x'\right)\right)^2}{2l^2}\right)$}, $\sigma^2 = 1$, $l=1$. $g(x) = \hat{g}(x) - \int \hat{g}(z) dz$.   & $\hat{\rho_0}(x) \sim \mathcal{GP}(0, k_p(x,x'))$, where $\sigma^2 = 1$, $l=1$, $\tilde{\rho_0}(x) = \text{softplus}\left(\hat{\rho_0}\left(x\right)\right)$, then we normalize the density function  to get $\rho_0(x) = \frac{\tilde{\rho_0}(x)}{\int \tilde{\rho_0}(z) dz}$.\\
    \hline
     18-19 &  see $\rho_0(x)$ for \# 15-17 & see $g(x)$ for \# 15-17.\\
    \hline
  \end{tabular}
    \caption{Operator, condition, and QoI parameters for differential equation problems listed in Table~\ref{tab:problem_table}.}
  \label{tab:problem_para}
\end{table}

For ODE-type of problems, we prepare $50$ equidistant key-value pairs for each $u$ and $c$ in the dataset, then select the first $n-1$ key-value pairs for $c$, and the first $n$ key-value pairs for $u$ when building the prompt, queries, and ground truth, with $n$ randomly sampled from 41 to 50. 

For the damped oscillator and PDE-type problems, we prepare $100$ equidistant key-value pairs for each condition/QoI in the dataset, then randomly select $n$ key-value pairs for each condition/QoI when building the prompt, queries, and ground truth, with $n$ randomly sampled from 41 to 50. 

For MFC-type problems, we prepare $100$ equidistant key-value pairs for each $g$, and prepare the key-value pairs for the density $\rho$ on a $51\times100$ uniform grid over the whole temporal-spatial domain, with $25\times100$ for $(t,x)\in[0.0.5)\times[0,1]$, and $26\times100$ for $(t,x)\in[0.5,1]\times[0,1]$. We randomly sample $n$ key-value pairs for each condition/QoI when building the prompt, queries, and ground truth, with $n$ randomly sampled from 41 to 50.

\bibliographystyle{unsrt}
\bibliography{biblist}

\end{document}